\documentclass[lettersize,journal]{IEEEtran}
\usepackage{amsmath,amsfonts}
\usepackage{algorithmic}
\usepackage{algorithm}
\usepackage{array}
\usepackage[caption=false,font=normalsize,labelfont=sf,textfont=sf]{subfig}
\usepackage{textcomp}
\usepackage{stfloats}
\usepackage{url}
\usepackage{verbatim}
\usepackage{graphicx}
\usepackage{cite}
\usepackage{comment}
\usepackage{multirow}
\usepackage{subcaption}
\usepackage{lipsum}
\usepackage[table]{xcolor}
\usepackage[table]{xcolor}
\usepackage{colortbl}
\usepackage{booktabs}
\usepackage{makecell}

\usepackage{relsize}
\usepackage{cellspace} 
\usepackage{url}            % simple URL typesetting
\usepackage{booktabs}       % professional-quality tables
\usepackage{nicefrac}       % compact symbols for 1/2, etc.
\usepackage{microtype}      % microtypography

\usepackage[dvipsnames]{xcolor}

\usepackage{tikz}
\usepackage{tikzscale} % To use includegraphics with tikz
\usetikzlibrary{shadows.blur}
\usepackage{tabularx} % X columns
\usepackage{multirow}
\usepackage{colortbl}
\usepackage{amsmath,amssymb,amsfonts}
\usepackage{diagbox} % diagonal corner tables
\usepackage{caption,subcaption} % subfigures
\usepackage{paralist}
\usepackage{glossaries}
\usepackage{xspace}

\usepackage{pifont}

\definecolor{orange}{RGB}{255,100,0}  % #ffe5cc
\definecolor{lred}{RGB}{255,204,204} %#ffcccc
\definecolor{lgreen}{RGB}{200,240,200} %#c8f0c8
\definecolor{lblue}{RGB}{204,229,255} %#cce5ff

\definecolor{darkbrown}{rgb}{0.4, 0.26, 0.13}
\definecolor{amethyst}{rgb}{0.6, 0.4, 0.8}
\definecolor{blue-violet}{rgb}{0.54, 0.17, 0.89}
\definecolor{caputmortuum}{rgb}{0.35, 0.15, 0.13}
\definecolor{darkviolet}{rgb}{0.58, 0.0, 0.83}
\definecolor{lavender}{rgb}{0.9, 0.9, 0.98}
\definecolor{indigo}{rgb}{0.29, 0.0, 0.51}

\newcommand{\steph}[1]{{\color{black}{#1}}}
\newcommand{\ours}{MultiGraspNet\xspace}
\newcommand{\mg}{multi-gripper\xspace}
\newcommand{\Mg}{Multi-gripper\xspace}
\newcommand{\mt}{multitask\xspace}
\newcommand{\Mt}{Multitask\xspace}

\usepackage{enumitem}
\setlist[itemize]{leftmargin=*}

\begin{document}

\title{\ours: A \Mt 3D Vision Model\\ for \Mg Robotic Grasping}

% \IEEEpubid{\vspace{4cm}
% This work has been submitted to the IEEE for possible publication. Copyright may be transferred without notice, after which this version may no longer be accessible.}
% \IEEEpubidadjcol

%\author{~\IEEEmembership{Anonymous Authors}

\author{
 Stephany Ortuno-Chanelo\textsuperscript{1},
 Paolo Rabino\textsuperscript{1},
 Enrico Civitelli\textsuperscript{2},
 Tatiana Tommasi\textsuperscript{1},
 Raffaello Camoriano\textsuperscript{1,3}

\thanks{Manuscript received: January 30, 2026; Accepted: June 11, 2026. This paper was recommended for publication by Editor Julia Borras Sol upon evaluation of the Associate Editor and Reviewers' comments.} 

 \thanks{This work was supported by Comau S.p.A. The authors would like to thank Giovanni Di Stefano, Simone Panicucci, Nicola Longo, Luca Di Ruscio, Luca Robbiano, and Simone Peirone for their valuable assistance. This study was carried out within the FAIR - Future Artificial Intelligence Research and received funding from the European Union Next-GenerationEU (PIANO NAZIONALE DI RIPRESA E RESILIENZA (PNRR) – MISSIONE 4 COMPONENTE 2, INVESTIMENTO 1.3 – D.D. 1555 11/10/2022, PE00000013). This manuscript reflects only the authors’ views and opinions, neither the European Union nor the European Commission can be considered responsible for them.}
 \thanks{
 \textsuperscript{1}VANDAL Laboratory, Department of Control and Computer Engineering, Politecnico di Torino, Turin, Italy.
 {\tt\footnotesize stephany.ortuno@polito.it}}
 \thanks{ 
 \textsuperscript{2}Comau S.p.A., Advanced Automation Solutions, Grugliasco, Italy.}
 \thanks{
 \textsuperscript{3}Istituto Italiano di Tecnologia, Genoa, Italy.
 }
}

        % <-this % stops a space
% <-this % stops a space
%\thanks{Manuscript received XXX; revised XX.}}

%---------------------For C version---------------------------
%Authors:Stephany Ortuno-Chanelo (1), Paolo Rabino (1), Enrico Civitelli (2), Tatiana Tommasi (1), Raffaello Camoriano (1)(3)
%Affiliations:(1) VANDAL Laboratory, Department of Control and Computer Engineering, Politecnico di Torino, Turin, Italy
%(2) Comau S.p.A., Advanced Automation Solutions, Grugliasco, Italy 
%(3) Istituto Italiano di Tecnologia, Genoa, Italy
%Aknowlegments:
%This work was supported by Comau S.p.A. The authors would like to thank Giovanni Di Stefano and Nicola Longo for their valuable assistance. This study was carried out within the FAIR - Future Artificial Intelligence Research and received funding from the European Union Next-GenerationEU (PIANO NAZIONALE DI RIPRESA E RESILIENZA (PNRR) – MISSIONE 4 COMPONENTE 2, INVESTIMENTO 1.3 – D.D. 1555 11/10/2022, PE00000013). This manuscript reflects only the authors’ views and opinions, neither the European Union nor the European Commission can be considered responsible for them. We thank the anonymous reviewers and editor for their constructive comments and suggestions.
%-----------------------------------------------------------------------------

\markboth{IEEE Robotics and Automation Letters. Preprint Version. Accepted June 2026}{Ortuno \MakeLowercase{\textit{et al.}}: MultiGraspNet} 


\maketitle

\begin{abstract}
Vision-based models for robotic grasping automate critical, repetitive, and draining industrial tasks. Existing approaches are typically limited in two ways: they either target a single gripper and are potentially applied on costly dual-arm setups, or rely on custom hybrid grippers that require ad-hoc learning procedures with logic that cannot be transferred across tasks, restricting their general applicability. 
In this work, we present \emph{\ours}, a novel \mt 3D deep learning method that predicts feasible poses simultaneously for parallel and vacuum grippers within a unified framework, enabling a single robot to handle both end-effectors. The model is trained on the richly annotated GraspNet-1Billion and SuctionNet-1Billion datasets, which have been aligned for the purpose, and generates graspability masks quantifying the suitability of each scene point for successful grasps.  By sharing early-stage features while maintaining gripper-specific refiners, \ours effectively leverages complementary information across grasping modalities. 
This design preserves a compact architectural footprint of only 15.75M parameters and enables fast inference on a single GPU, enhancing adaptability and efficiency in cluttered scenes.
We characterize \ours's performance with an extensive experimental analysis, demonstrating its competitiveness with single-task models on relevant benchmarks while reducing computational cost. Moreover, real-world experiments on a single-arm multi-gripper robotic setup show that our approach outperforms normalization-based \mg approaches.\\
\steph{Project page: \url{https://vandal-lab.github.io/multigraspnet-project}}

\end{abstract}


\section{Introduction}

\IEEEPARstart{I}{n} recent years, the demand for robotic systems in industrial and warehouse environments has grown significantly, expanding toward increasingly diverse and complex tasks. This trend is driven by the need to improve operational efficiency and working conditions, while reducing safety risks for human operators. A central challenge in this context is the development of robotic systems capable of handling a wide variety of items that differ in shape, size, weight, and material properties. In response, research in autonomous grasping has led to increasingly sophisticated solutions that integrate computer vision, machine learning, and robotics to perceive, localize, and manipulate objects in cluttered environments.

\begin{figure}[t]
\centering
\includegraphics[width=0.75\columnwidth]{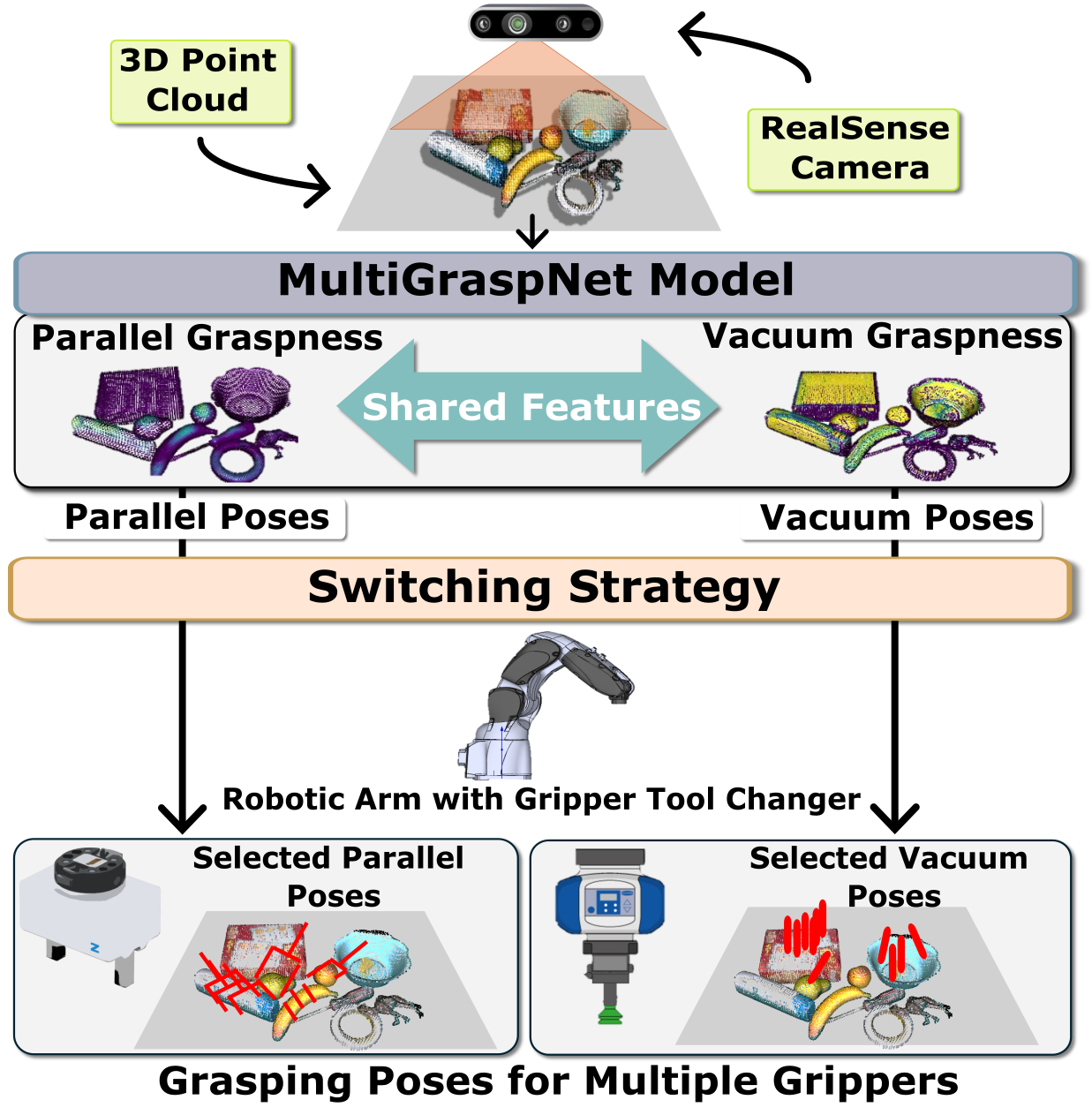}
\caption{Our proposed \mg approach processes a 3D scene point cloud and learns a shared representation to jointly predict grasping quality scores for parallel and vacuum grippers based on geometric cues. The resulting graspness maps are further refined to produce the final grasp poses for both grippers. Finally, a switching strategy module selects the optimal gripper type for execution without altering the predicted poses.}
\vspace{-5mm}
\label{fig:VA}
\end{figure}

Classical robotic grasping systems employ a single gripper, most commonly either a parallel gripper or a vacuum gripper. 
The former relies on frictional contacts and is best suited to rigid objects that provide sufficient friction and stable contact geometry for force closure~\cite{dex_par,wu2024economic}. 
The latter is most effective for objects with smooth, low-porosity surfaces that can form a seal~\cite{dex_vacuum,cao2024uncertainty,10711021}.
Although using one end-effector is cost-effective and simplifies grasp planning, performance is limited by the mechanical and geometric properties of the selected gripper. Thus, single-gripper robots often struggle to reliably grasp a diverse range of objects, reducing their adaptability and effectiveness.
These limitations have been addressed through the development of custom hybrid grippers 
combining suction cups~\cite{zeng2022robotic}, compliant fingers~\cite{zhu2022soft}, or sensor-integrated fingertips~\cite{burgess2025grasp, hu2024learning}. Such grippers are typically tailored to specific object characteristics, achieving high performance in targeted applications, but exhibiting limited adaptability when confronted with novel items or unanticipated configurations, which hinders generalization.

Recently designed systems overcome these issues by using  
different grippers with multiple robotic arms, working collaboratively to handle a broader range of objects~\cite{switching, olesen2020collaborative}. However, they present increased spatial requirements, financial costs, and overall complexity. 
Furthermore, despite improving grasping capabilities, these systems typically rely on separate, 
% \newpage %<------------------------------------------------- to avoid footpage
gripper-specific %
models, which can lead to suboptimal learning. This setup is computationally inefficient during both training and prediction, since each model is trained independently. Besides, the learned representations remain isolated, preventing %
knowledge transfer and limiting generalization to unseen scenarios.
In this context, \mg methods for a single robotic arm represent an underexplored, yet promising direction that can strike the balance between cost and performance. 
Crucially, the challenge goes beyond designing and controlling a robot capable of switching among multiple end-effectors, and calls for learning a multi-objective shared model within a single framework. \steph{Such model should build a unified internal representation that captures geometric features relevant to different grasping modalities.} 

For this purpose, we introduce our \emph{\ours}, a novel 3D vision-based \mt deep learning architecture that predicts grasping poses for two different grippers simultaneously. \steph{Overcoming the need for multiple specialized models, \ours leverages both early-stage shared representations and specialized pose refiners to extract tailored features for each gripper resulting in improved grasping success rate and a unified learning pipeline.} Our approach, outlined in Fig.~\ref{fig:VA}, allows a single robotic arm to utilize two distinct grippers: a parallel-jaw gripper and a vacuum gripper. This configuration enhances the versatility of the robot, allowing it to handle a significantly broader range of objects while reducing hardware and computational costs.

\smallskip
\noindent \textbf{To summarize, our main contributions are:} 
\begin{itemize}
    \item We introduce \ours, an 
    efficient 
    \mt 3D deep learning model capable of generating multiple candidate poses for vacuum and parallel grippers simultaneously from 3D visual input with a compact 15.75M parameter footprint.

    \item  We introduce a learning-based switching strategy that dynamically selects the most suitable gripper, effectively resolving the final execution choice between modality-specific grasp predictions.
    
    \item We evaluate our system with an extensive experimental analysis including benchmark comparisons, design-choice  
    and real-world grasping experiments on known and novel objects in an industrial scenario.

\end{itemize}
Our results demonstrate \ours's advantage over single-task baselines and highlight the potential of \mt learning for scalable and robust \mg robotic manipulation, paving the way for future research on unified grasping models and more adaptive robotic systems.

\section{RELATED WORKS}

\subsection{Suction Grasping}

Suction-based grasping is a particularly effective solution in cluttered environments due to its simplicity.
Early approaches to suction grasp prediction primarily relied on convolutional neural networks (CNNs)
to infer grasp poses from RGB or depth images, while alternative methods exploited purely geometric reasoning on point clouds: ~\cite{shao2019suction} proposed a hybrid ResNet-based U-Net architecture for dense suction grasp prediction.
SuctionNet-1Billion (S1B)~\cite{cao2021suctionnet} further advanced the field through a two-stage framework that integrates physics-based grasp evaluation and provides a large-scale benchmark for online evaluation.

Concerning 3D perception–based approaches, the Grasp Quality CNN (GQ-CNN)~\cite{dex_vacuum} method trained on the synthetic Dex-Net 3.0 dataset demonstrated strong generalization capabilities on novel objects. More recently, SGNet~\cite{zhai2025sgnet} showed how to predict suction grasp parameters and object centers directly from point cloud data, while UISN~\cite{cao2024uncertainty} proposed a multi-stage pipeline that focuses on unseen objects by using Unseen Object Instance Segmentation (UOIS)~\cite{xie2021unseen}.

Despite their effectiveness, these methods tend to perform best on relatively regular and flat objects, while being less reliable on highly textured or deformable objects, and on surfaces that hinder the formation of an airtight suction seal.

\vspace{-3 mm}

\subsection{Parallel Grasping}
Early approaches for parallel grasp detection 
introduced the grasping rectangle representation~\cite{jiang2011efficient}. 
Building on this idea,~\cite{lenz2015deep} proposed a cascaded two-network system that prunes unlikely grasps to balance speed and robustness. Subsequent works introduced region-based detectors combining global and local candidates to improve robustness and precision~\cite{asif2019densely}. More recently, RGBD-Grasp~\cite{gou2021rgb} decomposed parallel grasp prediction into two sub-tasks: inferring position and orientation, followed by grasp width and depth.

Among 3D perception-based methods, Dex-Net 2.0~\cite{dex_par} introduced a large-scale synthetic set of point clouds and grasping poses, with a first version of the GQ-CNN model that predicted parallel grasp success probabilities. Recognizing the need for larger datasets and better evaluation systems, GraspNet-1Billion~\cite{graspnet1b} provided a massive data collection 
from the real world with annotations obtained by analytic computation in simulation. 
To improve grasp candidate selection, GSNet~\cite{wang2021graspness} introduced the concept of \textit{graspness}, a geometry-driven quality measure, and developed a fast end-to-end network integrating it.
The authors of~\cite{alliegro2022end} introduced an end-to-end deep learning approach leveraging differentiable sampling. More recently, EconomicGrasp (EFG)~\cite{wu2024economic} tackled the challenge of costly supervision by efficiently selecting labels with low ambiguity. GenGrasp \cite{ma2024generalizing}, instead, explicitly modeled grasping physical priors to enhance novel object manipulation.

Although parallel-gripper approaches are well-suited for handling complex objects, they can face difficulties with large or flat objects of limited height. In such cases, incorporating an end-effector with complementary capabilities, such as a vacuum gripper, can enhance  
effectiveness.

\vspace{-3 mm}

\subsection{\Mg Grasping} 
To overcome the limitations of each end-effector, several studies focused on the adoption of multi-functional grippers. 
For instance, Carman~\cite{morrison2018cartman} won the Amazon Robotics Challenge in 2017 with a custom grasping tool that included suction and parallel grippers. Their model relied on a semantic segmentation network to predict grasping poses, followed by heuristic grasp synthesis. In the same challenge,
~\cite{zeng2022robotic} presented an approach for predicting affordance maps over four grasping primitives, enabling the use of a custom gripper that combines suction and parallel grasping.
Instead,~\cite{switching} proposed a robot system able to use two gripper combinations via a switching strategy. It exploited a Single Shot MultiBox Detector (SSD) network that includes objectness prediction, and the choice of the gripper is based on the sparseness of the identified object items. In~\cite{olesen2020collaborative}, a switching strategy was introduced for suction and multi-fingered grippers to perform random bin picking and assembly applications. Even in this case the learning component was limited to object recognition and localization.  

Dex-Net 4.0~\cite{dexnet4} trained separate GQ-CNNs to predict the probability of grasp success for parallel and suction grippers given a point cloud, rather than pursuing a unified model. 
In contrast, Corner-Grasp~\cite{son2025corner} introduced a \mt learning approach that infers 2D grasping poses for multiple grippers and predicts surface material to select the grasping modality. 

While prior work on \mg grasping began exploring the benefits of \mt learning for jointly modeling multiple grasping modalities, existing approaches are primarily validated on 2D data in synthetic environments or are constrained by custom hardware.

\steph{Building on this emerging direction,  \ours introduces a \mt framework that operates in 3D on real-world data, effectively unifying distinct grasping modalities by leveraging feature sharing within a single and efficient architecture.} 

\begin{figure}[t]
\centering
\includegraphics[width=0.5\columnwidth]{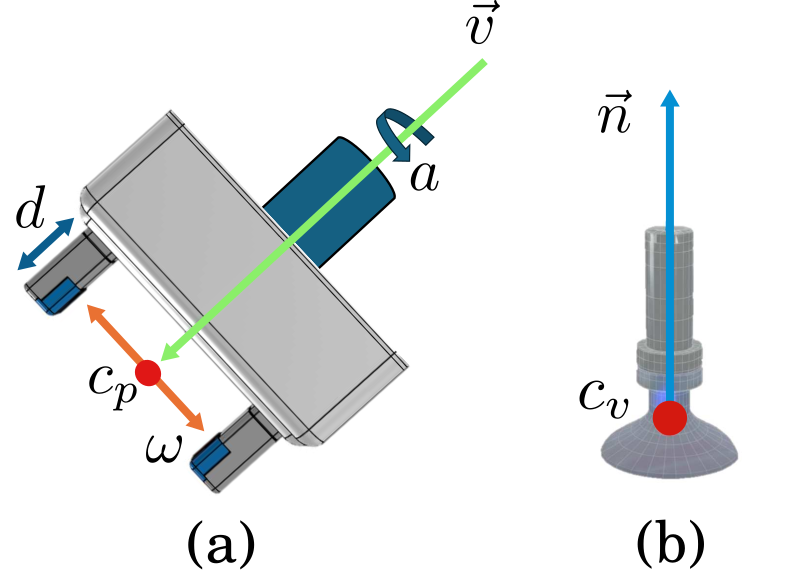}
  \caption{Schematic illustration of the parallel-jaw gripper (a) 
  and vacuum gripper (b) poses. The former includes the central point $c_p$, the depth $d$, the width $\omega$, as well as the approach direction $v$ and angle $a$. The vacuum grasp pose includes the central point $c_v$ and the normal to the point $n$. 
  } \vspace{-2mm}
\label{fig:gripper_definition}
\end{figure}

\section{Method}
\subsection{Task Definition}
We introduce a unified \mt deep learning framework that enables a single model to predict grasping poses for two heterogeneous end-effectors, 
as illustrated in Fig.~\ref{fig:gripper_definition}. \steph{Our approach is designed to increase the system’s versatility by leveraging the advantages of each gripper within a single framework capable of learning features relevant to both tasks.}

Given an input scene $\{p_i \in \mathbb{R}^3\}_{i = 1}^N$, formally represented as a 3D point cloud of size $N$, the goal of our model is to predict candidate grasping poses for two gripper types simultaneously.
Following the conventions established in \cite{wu2024economic}, we define the 3D grasping pose for the parallel-jaw gripper as: 
\begin{equation}
\label{parallel_pose}
G_p = [c_{p}, v, a , \omega, d, \steph{s_{p}}],
\end{equation}
where $c_{p}$ indicates the central point of the grasp in the 3D  
space, $v$ is the approaching direction vector and $a$ [$^{\circ}$] represents the angle of the 2D in-plane rotation. The remaining three components $d$, $\omega$, and \steph{$s_{p}$} are respectively the depth [m], which is the distance from the grasp point to the origin of the gripper frame, the width [m], that describes the distance between the fingers needed to grasp the object, and the grasp score which indicates the quality of the pose. 

For the vacuum-based gripper, the grasp pose is represented as in \cite{cao2021suctionnet}:
\begin{equation}
\label{parallel_pose}
G_v = [c_{v},n,\steph{s_{v}}],
\end{equation}
where $c_{v}$  
is the central grasp point in the 3D space, $n$ 
is an estimate of the 3D surface normal vector in that point, and \steph{$s_v$} is the vacuum score indicating the quality of the pose. 

\begin{figure*}[t]
    \centering
    \includegraphics[width=0.93\textwidth]{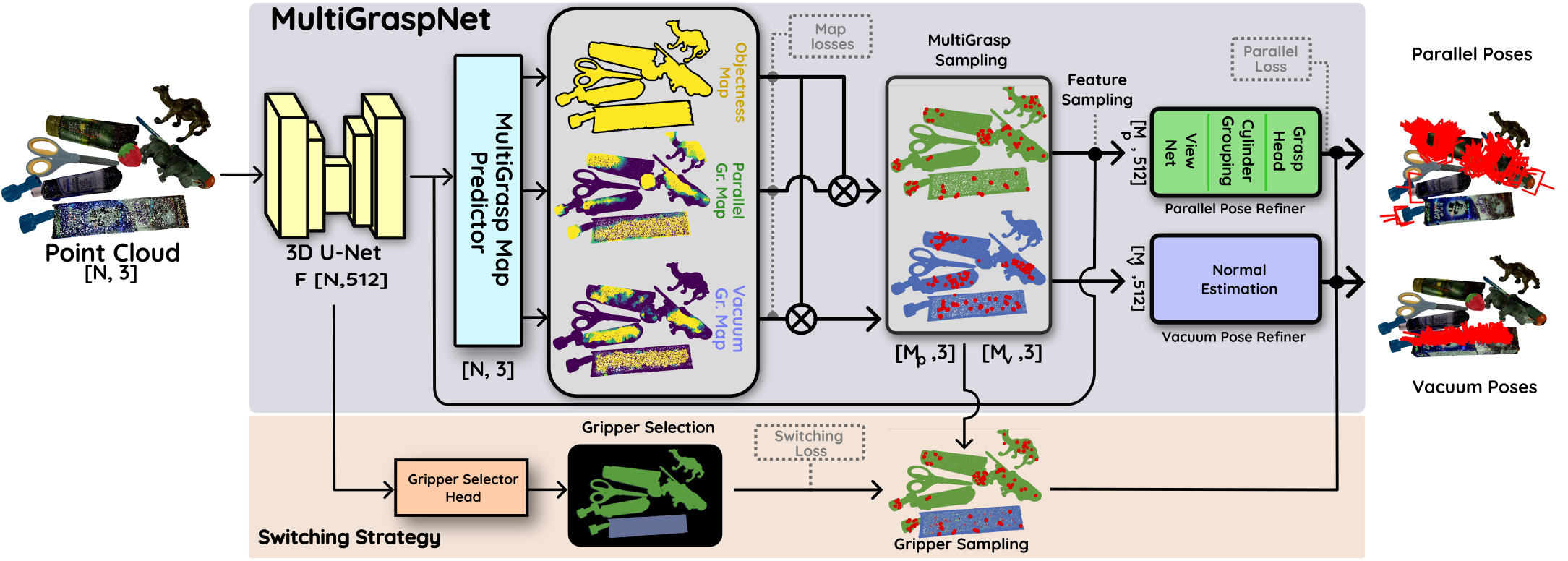}
    \caption{ %Overview of our architecture. 
    The network takes as input a 3D point cloud. Then the Minkowski-based backbone extracts geometric features and \steph{these are} processed by a multi-branch grasp prediction head. Gripper-specific refinement modules are applied to generate the \mg grasping poses. Finally, a Switching Strategy module predicts a mask to filter these poses by the optimal gripper, without modifying the underlying MultiGraspNet predictions.
    } \vspace{-2mm}
    \label{fig:architecture}
\end{figure*}

\vspace{-1.5 mm}

\subsection{Dataset with \Mg Graspness Map}

To obtain a dataset
with \mg annotations,
we \steph{align}

the GraspNet-1Billion~\cite{graspnet1b} and SuctionNet-1Billion~\cite{shao2019suction} datasets, which share a common set of 
scenes. 

We define a \emph{graspness score map} that encodes the suitability of scene regions for grasping across two end-effectors, enabling multi-task training and fair comparison with single-task methods.
For the parallel gripper, the map is computed by starting from the provided friction coefficient and using the look-ahead search strategy for grasp success introduced in \cite{wang2021graspness}. 
In this letter, this strategy is extended to the vacuum task over single objects 
by considering the provided seal coefficient and filtering out unsuccessful grasps (seal coeff $<0.004$). Then, the graspness map is obtained at the scene level by using the 6D pose for each object and filtering out the poses with collisions. Finally, the map is projected onto the original point cloud.

To exclude non-graspable locations, points outside the operational zone (table depth $<$ 0) are pruned, together with background points not belonging to any object instance using the \textit{objectness} score $o$. Then, a nearest neighbor search associates each scene point with its closest grasp properties.
The per-point vacuum graspness map is rescaled to the $[0,1]$ range, after which low-quality grasp regions are filtered out (scores below 0.1).

Overall, a supervised \mg graspness map with the needed \steph{$r_p$} and \steph{$r_v$} values is constructed, derived from physically accurate grasping poses and accounting for the force-closure coefficient in the case of the parallel gripper and the seal coefficient for the vacuum gripper. This formulation enables the model to learn meaningful graspability patterns grounded in physically valid grasp configurations.

\subsection{\ours}
Our newly designed \mg model is presented in Fig.~\ref{fig:architecture}. The rest of this section describes its main components. 

\subsubsection{Backbone Network}
Similar to previous approaches~\cite{wang2021graspness,wu2024economic}, we use a 3D U-Net as the main backbone, which is based on MinkowskiEngine \cite{choy20194d-minkowski}. The backbone encodes geometric scene information into per-point 
vectors, providing as output a feature tensor of shape $[N,512]$. 
\subsubsection{MultiGrasp Map Predictor}
This module consists of three heads that elaborate in parallel on the features to output, respectively, the 
objectness score $\hat{o}$, the graspness scores for the parallel gripper \steph{$\hat{r}_p$}, and the graspness scores for the vacuum gripper \steph{$\hat{r}_v$}. Every head is defined by a 1D convolutional layer that independently predicts a score for each point, obtaining a $[N,3]$-dimensional output that can be visualized onto the scene as the graspness and objectness maps.
During training, the objectness head is optimized using a standard cross-entropy loss $\mathcal{L}_{obj}$ as in \cite{wu2024economic}, while both graspness heads are trained using a weighted binary cross entropy loss:
\begin{equation}
    \mathcal{L}_{graspness} = - \frac{1}{N} \sum_{i=1}^{N} [\lambda \cdot \steph{r_i} \log(\steph{\hat{r}_{i}}) + (1 - \steph{r_i}) \log(1 - \steph{\hat{r}_{i}})].
    \label{graspness_eq}
\end{equation}
For the parallel graspness loss $\mathcal{L}_{graspness}^{par}$, we set the positive samples weight $\lambda = 10$ to account for graspable points sparsity, while  $\lambda = 1$ for the vacuum graspness loss $\mathcal{L}_{graspness}^{vac}$.

\subsubsection{MultiGrasp Sampling}
Combining the task-specific graspness maps with objectness enables the identification of object points with a high probability of successful grasping. We formalize this via per-point score pair multiplication (denoted by $\otimes$ in Fig.~\ref{fig:architecture}), followed by thresholding ($t_p$=$t_v$=0.1) to discard low-scoring points. 
To ensure maximum spatial coverage, the points are sub-selected via Farthest Point Sampling (FPS). Finally, 1024 seed points for each gripper ($M_p$, $M_v$), paired with their 512-dimensional backbone features, are passed to the respective parallel and vacuum pose refiners.

\subsubsection{Parallel Pose Refiner}

Following \cite{graspnet1b} and \cite{wang2021graspness}, the $M_p$ seeds are fed to \textit{ViewNet} and \textit{Cylinder Grouping} to determine approaching vectors $v$. The training process optimizes a Smooth L1 loss for view ($\mathcal{L}_{view}$) and width prediction ($\mathcal{L}_{width}$), while a cross-entropy loss is adopted for the in-plane angle ($\mathcal{L}_{angle}$) and depth ($\mathcal{L}_{depth}$). 
Meanwhile, the grasp score \steph{$s_{p}$} estimation is formulated as a classification task and supervised via the cross-entropy loss $\mathcal{L}_{score}$.

\subsubsection{Vacuum Pose Refiner}
The vacuum pose prediction follows a different strategy due to the simpler nature of the pose. The refiner does not need further losses and is only adopted at inference time (i.e., \steph{$s_v=\hat{r}_v$}). 
For each of the points in $M_v$, the surface normal vector $n$ is estimated by looking at the geometry of its nearby neighbors. Using an off-the-shelf normal estimation method~\cite{zhou2018open3d}, a small radius is used to find the closest points around each seed, and the normal vector is computed by analyzing the local shape through covariance to fit the best local surface. These normals complete the vacuum poses directly from point geometry without requiring additional view selection or grouping operations. 

Overall, the design of our \mg model combines the advantages of feature sharing and task-specific refiners to handle the unique characteristics of different end-effectors, enabling multi-grasp predictions within a single unified framework, avoiding the use of costly separate architectures.

The entire network is trained end-to-end. 
The total loss $\mathcal{L}_{total}$ for \ours is formulated 
by combining the individual task-specific losses as:

\begin{equation}
\begin{aligned}
    \mathcal{L}_{total} = & ~\mathcal{L}_{obj} + \alpha\mathcal{L}_{graspness}^{vac} + \beta\mathcal{L}_{graspness}^{par} + \gamma\mathcal{L}_{view}  \\
   +& ~\mathcal{L}_{angle} + \mathcal{L}_{depth} + \mathcal{L}_{score} + \delta\mathcal{L}_{width}~,
\end{aligned}
\label{loss_eq}
\end{equation}
where ad-hoc weights are introduced and calibrated to maintain comparable scales across the loss terms. 
We set the view selection weight as $\gamma = 100$, while the weights for the width prediction, parallel, and vacuum grippers are respectively set as $\delta=\alpha=\beta=10$.

\subsubsection{Switching Strategy}

A key challenge in \mg grasp prediction is that the grasp quality scores produced for different end-effectors are not directly comparable, as \steph{$s_p$} and \steph{$s_v$} are derived from fundamentally different physical principles.
Thus, a naïve comparison between these scores may lead to ambiguous or suboptimal gripper selection when both types yield high confidence on the same object area.
To address this issue, we introduce a dedicated \emph{Gripper Selector Head} that determines the most suitable gripper type for each object based on its geometric properties. Concretely, we implement an additional head that operates on the pretrained 3D U-Net backbone features. This module predicts a per-point probability vector $\hat{y}\in(0,1)^3$, whose entries correspond to the background, parallel, or vacuum gripper. The one-hot encoded  ground-truth vector $y\in\{0,1\}^3$ for this head is annotated based on geometric considerations.
\steph{Specifically, labels are assigned by expert human operators  primarily evaluating local surface flatness.\footnote{\steph{G1B\cite{graspnet1b} object IDs assigned to vacuum: $\{1, 2, 7, 36, 40, 47, 59, 69\}$. All the remaining ones are assigned to parallel.}}} 
During training, this head is optimized using a multi-class cross-entropy loss:
\begin{equation}
    \mathcal{L}_{switching} = - \frac{1}{N}\sum_{i=1}^N \sum_{k=1}^{3} y_{k,i} \log(\hat{y}_{k, i}).
\end{equation}
At inference time, the predicted vector is used to filter the grasp candidates generated by \ours. In particular, for each object, only the grasp poses corresponding to the selected gripper type are retained. 
This enforces a consistent assignment of a single gripper per object, preventing conflicting grasp proposals.

\vspace{-2mm}

\subsection{Implementation Details} \label{ImpDet}
Our approach addresses two tasks corresponding to two end-effectors (vacuum and parallel).  
To efficiently integrate these tasks,  we adopt the Gradient Surgery for \Mt Learning (PCGrad) technique \cite{yu2020gradient} during training, which mitigates conflicting gradient updates across tasks.
For training, we use the Adam optimizer with an initial learning rate of $5\times10^{-4}$ and a cosine decay scheduler.
The batch size is set to 12, and training runs for 22 epochs.  
All experiments are conducted using PyTorch on a single NVIDIA GeForce RTX 5090 GPU.
\section{EXPERIMENTS}

\subsection{Experimental Protocol and Metrics}
The defined dual-gripper dataset comprises 190 scenes, of which 100 are used for training and 90 for testing. Following~\cite{cao2021suctionnet, graspnet1b}, we divide the test set into three groups: 
seen/similar/novel. This enables us to evaluate how different methods generalize in increasingly complex scenarios. 
The grasp prediction modules generate multiple grasp candidates for each scene.
Therefore, by following standard practice \cite{graspnet1b, cao2021suctionnet}, we adopt Precision@$k$ as our evaluation metric, which measures precision among the top-k ranked grasps. 
We denote by $AP_{\mu_p}$ and $AP_{\mu_v}$ the Average Precision@$k$ for the parallel and vacuum tasks respectively, with $k$ ranging from 1 to 50. The subscripts $\mu_p$ and $\mu_v$ refer to specific friction (parallel) and seal (vacuum) coefficients. Note that a high $\mu_p$ indicates that the gripper applies a large force, which generally leads to a higher-quality grasp. For the vacuum gripper, a grasp is considered positive only when the seal coefficient exceeds the value indicated by the subscript. Therefore, a high $\mu_v$ corresponds to a more selective and stricter success criterion. 
We also report the overall AP as the average result with $\mu_p$ ranging from $0.2$ to $1.0$ and $\mu_v$ ranging from $0.2$ to $0.8$, both with an interval of $0.2$.
\begin{table}[t]
    \centering
    \caption{Single-task results on the RealSense data of the aligned dual-gripper dataset from SuctionNet-1Billion (Vacuum) and GraspNet-1Billion (Parallel). Top results are in bold, while the second best are underlined.} 
    \label{table1:singletask}

    \setlength\extrarowheight{2pt}

    \resizebox{\columnwidth}{!}{    
    \begin{tabular}{l@{~~}c@{~~}c@{~~}c@{~~}c@{~~}c@{~~}c@{~~}c@{~~}c@{~~}c}
        \toprule
        \multicolumn{10}{c}{\textbf{Vacuum}}\\
        \hline
        {} & \multicolumn{3}{c}{Seen} & \multicolumn{3}{c}{Similar} & \multicolumn{3}{c}{Novel} \\
        \cmidrule(lr){2-4} \cmidrule(lr){5-7} \cmidrule(lr){8-10}
        Method & $AP$ & $AP_{0.8}$ & $AP_{0.4}$ & $AP$ & $AP_{0.8}$ & $AP_{0.4}$ & $AP$ & $AP_{0.8}$ & $AP_{0.4}$ \\
        \midrule
        DexNet3.0 \cite{dex_vacuum}  & 15.50 & 1.53 & 20.22 & 18.92 & 2.62 & 24.51 & 2.62  & 0.35  & 5.32 \\
        S1B \cite{cao2021suctionnet} & \textbf{28.31} & 3.41 & \textbf{38.56}  & 26.64  & 3.42 & 35.34 & \textbf{8.23} & 0.35  & \textbf{10.29} \\
        \rowcolor{gray!15}
        \ours-vac.& \underline{28.26} & \underline{3.60} & \underline{36.39}
        & \textbf{31.37} & \textbf{3.88} & \underline{41.63}
        & \underline{8.07} & \textbf{0.48} & \underline{10.03} \\
        \rowcolor{gray!15}
        \ours & 28.07 & \textbf{3.62} & 35.81 & \underline{31.22} & \underline{3.53} & \textbf{41.65} & 7.38 & \underline{0.46} & 8.92 \\
        \midrule
        \multicolumn{10}{c}{\textbf{Parallel}}\\
        \hline
        {} & \multicolumn{3}{c}{Seen} & \multicolumn{3}{c}{Similar} & \multicolumn{3}{c}{Novel} \\
        \cmidrule(lr){2-4} \cmidrule(lr){5-7} \cmidrule(lr){8-10}
        Method & $AP$ & $AP_{0.8}$ & $AP_{0.4}$ & $AP$ & $AP_{0.8}$ & $AP_{0.4}$ & $AP$ & $AP_{0.8}$ & $AP_{0.4}$ \\
        \midrule
        G1B \cite{graspnet1b}  & 27.56 & 33.43 & 16.95 & 26.11 & 34.18 & 14.23 & 10.55 & 11.25 & 3.98 \\
        GSNet \cite{wang2021graspness} & 67.12 & 78.46 & 60.90 & 54.81 & 66.72 & 46.17 & 24.31 & 30.52 & \textbf{14.32} \\
        EFG \cite{wu2024economic} & \underline{68.21} & \underline{79.60} & \underline{63.54} & \textbf{61.19} & \textbf{73.60} & \underline{53.77} & \underline{25.48} & \underline{31.46} & \underline{13.85} \\
        \rowcolor{gray!15}
        \ours-par. & 67.70 & 78.97 & 62.67
        & \underline{60.61} & 72.51 & \textbf{54.01}
        & 25.45 & 31.36 & 13.60 \\
        \rowcolor{gray!15}
        \ours & \textbf{68.37} & \textbf{79.61} & \textbf{64.02} & 60.28 & \underline{72.54} & 52.87 & \textbf{25.53} & \textbf{31.67} & 13.66 \\
        \bottomrule
    \end{tabular}
    } 
    % \vspace{-3mm}
\end{table}

\begin{table}[t]
\centering
\caption{Small-Data Regime Analysis}
\label{table:sdr}
\resizebox{\columnwidth}{!}{%
\begin{tabular}{llcccccc}
\toprule
& & \multicolumn{6}{c}{\textbf{Number of scenes}} \\
\cmidrule(lr){3-8}
\textbf{Metric} & \textbf{Model} & \textbf{1} & \textbf{2} & \textbf{3} & \textbf{4} & \textbf{5} & \textbf{6} \\
\midrule
\multirow{2}{*}{$\overline{AP_{v}}$} & \ours-vac. & $15.19$ & $16.66$ & $17.59$ & $18.26$ & $18.17$ & $18.85$ \\
                                     & \ours     & $15.14$ & $17.17$ & $18.12$ & $17.80$ & $18.71$ & $19.15$ \\
\midrule
\multirow{2}{*}{$\overline{AP_{p}}$} & \ours-par. & $15.30$ & $20.73$ & $25.32$ & $29.33$ & $32.74$ & $33.59$ \\
                                     & \ours     & $14.48$ & $20.38$ & $24.16$ & $28.95$ & $32.18$ & $33.93$ \\
\bottomrule
\end{tabular}%
}
\vspace{-3.5mm}
\end{table}

\subsection{Benchmark Comparison}
\steph{In this first analysis, we are interested in validating the core functioning of our model, so we report the performance of its single-task variants, either vacuum (\emph{\ours-vac.}) or parallel (\emph{\ours-par.}), obtained by deactivating the objective of the other gripper during training. The \mt version optimizes the loss in eq. (\ref{loss_eq}), thus it does not include the switching strategy.} 
We compare the results of our \ours with well-established single-gripper approaches, including S1B \cite{cao2021suctionnet}, Dex-Net3.0 \cite{dex_vacuum} for the vacuum, and G1B \cite{graspnet1b}, GSNet \cite{asif2019densely}, EFG~\cite{wu2024economic} for the parallel end-effector. 

\steph{The upper section of Table \ref{table1:singletask}} presents the vacuum grasping results. For $AP_{0.8}$, our vacuum-specific variant \ours-vac. 
consistently outperforms competing methods across all object groups. This advantage is particularly pronounced for similar objects, where improvements are also observed for $AP_{0.4}$ and $AP$. The full \mt version \ours preserves the gain in most of the considered cases, with minor positive or negative fluctuations. 

The bottom portion of the table reports the results for parallel grasping. All competing methods in this setting are 3D point-cloud–based and have model capacities comparable to ours. While the single-task variant \ours-par. 
achieves slightly lower performance than the best EFG competitor with a gap of less than one $AP$ point, the full \mt \ours attains the best results on both the seen and novel object groups.

\steph{We also consider a sub-sampled training set consisting of one to six scenes. We report the average AP across splits (seen, similar, novel) for vacuum $\overline{AP_{v}}$ and parallel $\overline{AP_{p}}$ tasks in Table  \ref{table:sdr}. The results confirm that \ours maintains competitive performance to the single-task variants across both modalities even in small data regimes.}

\begin{figure*}[t]
\centering
% \vspace{-1mm}
\includegraphics[width=1\textwidth]{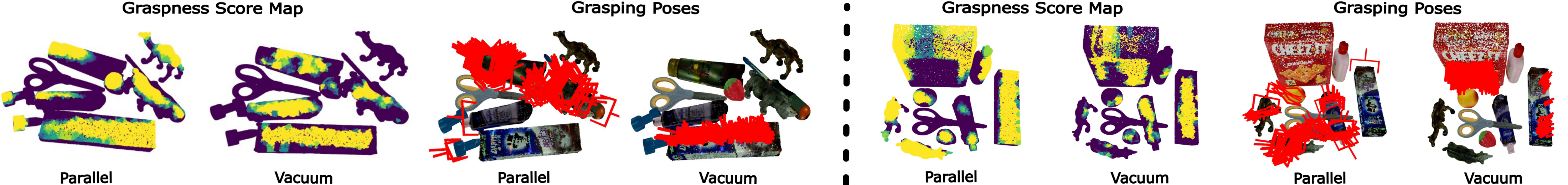} 
% \vspace{-2mm}
  \caption{Qualitative results of our \ours for two different scenes. 
  Our model outputs tailored graspness score maps 
  for each end-effector, which then guide the prediction of the grasping poses in an end-to-end framework. The shown grasping poses are the top 100 obtained for each 
  gripper after the filtering provided by the switching strategy module.}
  %\vspace{-3mm}
\label{fig:prediction}
\end{figure*}

Fig. \ref{fig:prediction} presents qualitative results obtained by \ours. 
Our model successfully generates feasible grasps for both grippers in diverse scenes.
Note that the predicted vacuum poses primarily concentrate on flat and smooth surfaces, such as boxes or planar regions (e.g., Cheez it box).
Instead, parallel grasp predictions focus on smaller and more complex objects (e.g., screws).
This complementary behavior highlights \ours's ability to adapt to the distinct properties of different grippers. 

\vspace{-2.5 mm}

\subsection{Exploring Design Variants}
Here, we provide a detailed analysis of how key design choices of \ours contribute to its performance.
We empirically show that employing gradient surgery in \mt training and excluding color features yields overall quantitative advantages across tasks.

\begin{table*}[t]
    \centering
    \caption{Analysis of design variants for \ours. 
    Results on the RealSense data of the aligned dual-gripper dataset from SuctionNet-1Billion (Vacuum) and GraspNet-1Billion (Parallel). Top results are in bold, while the second best are underlined.}
    \label{table:combined}

    \setlength\extrarowheight{2pt}

    \resizebox{\textwidth}{!}{
    \begin{tabular}{l@{~~}c@{~~}c@{~~}c@{~~}c@{~~}c@{~~}c@{~~}c@{~~}c@{~~}c@{~~}|c@{~~}c@{~~}c@{~~}c@{~~}c@{~~}c@{~~}c@{~~}c@{~~}c}  
        \toprule
        \rowcolor{white}
        & 
        \multicolumn{9}{c|}{\textbf{Parallel}} 
        & \multicolumn{9}{c}{\textbf{Vacuum}} \\
        \cmidrule(lr){2-10} \cmidrule(lr){11-19}

        & 
        \multicolumn{3}{c}{Seen} &
        \multicolumn{3}{c}{Similar} &
        \multicolumn{3}{c|}{Novel} &
        \multicolumn{3}{c}{Seen} &
        \multicolumn{3}{c}{Similar} &
        \multicolumn{3}{c}{Novel} \\
        \cmidrule(lr){2-4}
        \cmidrule(lr){5-7}
        \cmidrule(lr){8-10}
        \cmidrule(lr){11-13}
        \cmidrule(lr){14-16}
        \cmidrule(lr){17-19}

        Model  
        & AP & $AP_{0.8}$ & $AP_{0.4}$
        & AP & $AP_{0.8}$ & $AP_{0.4}$
        & AP & $AP_{0.8}$ & $AP_{0.4}$
        & AP & $AP_{0.8}$ & $AP_{0.4}$
        & AP & $AP_{0.8}$ & $AP_{0.4}$
        & AP & $AP_{0.8}$ & $AP_{0.4}$ \\
        \midrule

         Corner-Grasp \cite{son2025corner} backbone& 66.36  & 77.16 & 61.88
        & 45.32 & 55.39 & 35.92
        & 21.17 & 26.47& 11.33
        & 26.74&  3.31& 34.91
        & 28.96 &  \textbf{3.69}& 35.92
        & \textbf{9.34} &  0.43& \textbf{11.33}\\

        \ours w/o PCGrad & \textbf{69.18} & \textbf{80.00} & \textbf{65.13}
        & \underline{60.08} & \textbf{72.59} & \underline{52.33}
        & \underline{24.92} & \underline{30.75} & \underline{13.41}
        & 26.70 & 2.90 & 34.60
        & 28.54 & 2.86 & 37.02
        &\underline{8.64} & \textbf{0.49} & \underline{10.56} \\

        \ours w/ RGB & 67.09 & 78.19 & 62.07
        & 58.49 & 70.60 & 50.73
        & 24.10 & 29.80 & 12.86
        & \underline{27.22} & \underline{3.01} & \underline{35.22}
        & \underline{30.06} & 3.40 & \underline{39.35}
        & 7.57 & 0.41 & 9.40 \\

        \rowcolor{gray!15}
        \ours & \underline{68.37} & \underline{79.61} & \underline{64.02}
        & \textbf{60.28} & \underline{72.54} & \textbf{52.87}
        &\textbf{25.53} & \textbf{31.67} & \textbf{13.66}
        & \textbf{28.07} & \textbf{3.62} & \textbf{35.81}
        & \textbf{31.22} & \underline{3.53} & \textbf{41.65}
        & 7.38 & \underline{0.46} & 8.92 \\

        \bottomrule
    \end{tabular}}
    % \vspace{-5mm}
\end{table*}

\subsubsection{Gradient Surgery for \Mt Learning}
\mt learning often suffers from optimization challenges due to gradient interference between tasks, which can lead to imbalanced training and cause certain tasks to dominate the learning process. 
Here, we compare the performance of \ours trained with the Adam optimizer with and without (w/o) PCGrad. As shown in Table~\ref{table:combined}, the model without PCGrad achieves noticeable improvements in parallel/seen and vacuum/novel, albeit at the cost of decreased performance across other groups.  To provide a single summary metric for task performance, we average the $AP$ values across the three categories (seen, similar, and novel) for each gripper, and we name it $\overline{AP}$. 
The performance of our model using just the Adam optimizer results in $\overline{AP}_p=51.39$ for parallel and $\overline{AP}_v=21.29$ for vacuum.  
When incorporating PCGrad, \ours yields 
$\overline{AP}_p=51.39$ for parallel and $\overline{AP}_v=22.22$ for vacuum. 
We conclude that PCGrad \steph{leads} to a more balanced and consistent performance across grasping modalities and splits. Thus, we always train our model with PCGrad.

\vspace{-0.2 mm}

\subsubsection{Adding RGB features} 
To test whether additional color information could improve scene understanding, we extend our model to incorporate 2D RGB features extracted from a pretrained ResNet-18 backbone. 
They are concatenated with the 3D features obtained from the point cloud, allowing the network to leverage both geometric and appearance-based cues. 
As shown in Table~\ref{table:combined}, \ours with (w) RGB features underperforms across tasks and splits with a small advantage only on novel objects for the vacuum gripper. 
These findings indicate that, within our setup, adding color information does not provide a reliable benefit and may even introduce unnecessary complexity without improving grasping performance. Thus, we decided not include RGB data for training our model.

\vspace{-0.2 mm}

\subsubsection{Corner-Grasp Backbone}

To further assess the impact of the feature extraction module, we conduct an additional comparison with the Corner-Grasp~\cite{son2025corner}. %baseline.
Although the original implementation is not publicly available, we replicate its backbone by adopting a ResNet101+FPN configuration and integrating it into our framework as a replacement for the 3D U-Net backbone.
This modification allows us to isolate the contribution of the backbone while keeping the remaining components of our pipeline unchanged. 
The results in Table~\ref{table:combined} highlight the effectiveness of our original design, showing that the 3D U-Net backbone provides more suitable geometric features for grasp prediction. Leading to improved performance particularly for the parallel gripper, which requires richer 3D representations to handle more complex geometries. 
Specifically, our model outperforms \steph{CornerGrasp \cite{son2025corner}} baseline across all categories, achieving a significant margin of over 2.0 AP points in the parallel gripper tasks and showing competitive results in the vacuum grasping modality.
%\vspace{-4.5 mm}

%\vspace{-3.5 mm}

\subsection{Efficiency Analysis}
We quantitatively evaluate the computational efficiency of our approach by comparing it with existing methods in terms of model size, number of parameters, and runtime performance. 
As reported in Table~\ref{tab:efficiency_comparison}, our model achieves a favorable trade-off between accuracy and efficiency, requiring only $15.75~\times~10^{6}$ parameters and 60.09~MB memory, which is significantly lower than both Corner-Grasp and the combined S1B+EFG setup. Besides, our method maintains a competitive runtime performance, achieving a latency of 50.78~ms on a single GPU. 
These results demonstrate that \ours can effectively handle multi-gripper grasp prediction without incurring in substantial computational overhead, and is therefore well-suited for real-time robotic applications under resource constraints.

\begin{table}
\centering
\caption{Model efficiency comparison.}
\resizebox{\columnwidth}{!}{
\begin{tabular}{lcccc}
\toprule
 & \textbf{\# Parameters $\downarrow$} & \textbf{Size [MB]} $\downarrow$ & \textbf{FPS $\uparrow$} & \textbf{Latency [ms] $\downarrow$} \\
\midrule
S1B \cite{cao2021suctionnet} & $58.75 \times 10^{6}$ & 224.12 & 12.48 & 80.09  \\
EFG \cite{wu2024economic} & $15.75 \times 10^{6}$ & 60.08 & 21.97 & 45.50  \\
S1B+EFG & $74.50 \times 10^{6}$ & 284.20 & 7.91 & 125.59 \\
CornerGrasp \cite{son2025corner} & $108.97 \times 10^{6}$ & 415.70 & 19.50 & 51.27 \\\rowcolor{gray!15}
MultiGraspNet & $15.75 \times 10^{6}$ & 60.09 & 19.69 & 50.78 \\
\bottomrule
\end{tabular}
}
\label{tab:efficiency_comparison}
\end{table}

\vspace{-2.5 mm}
\subsection{Real-world Experiments}

\begin{figure}[t]
\centering
\includegraphics[width=1.0\columnwidth]{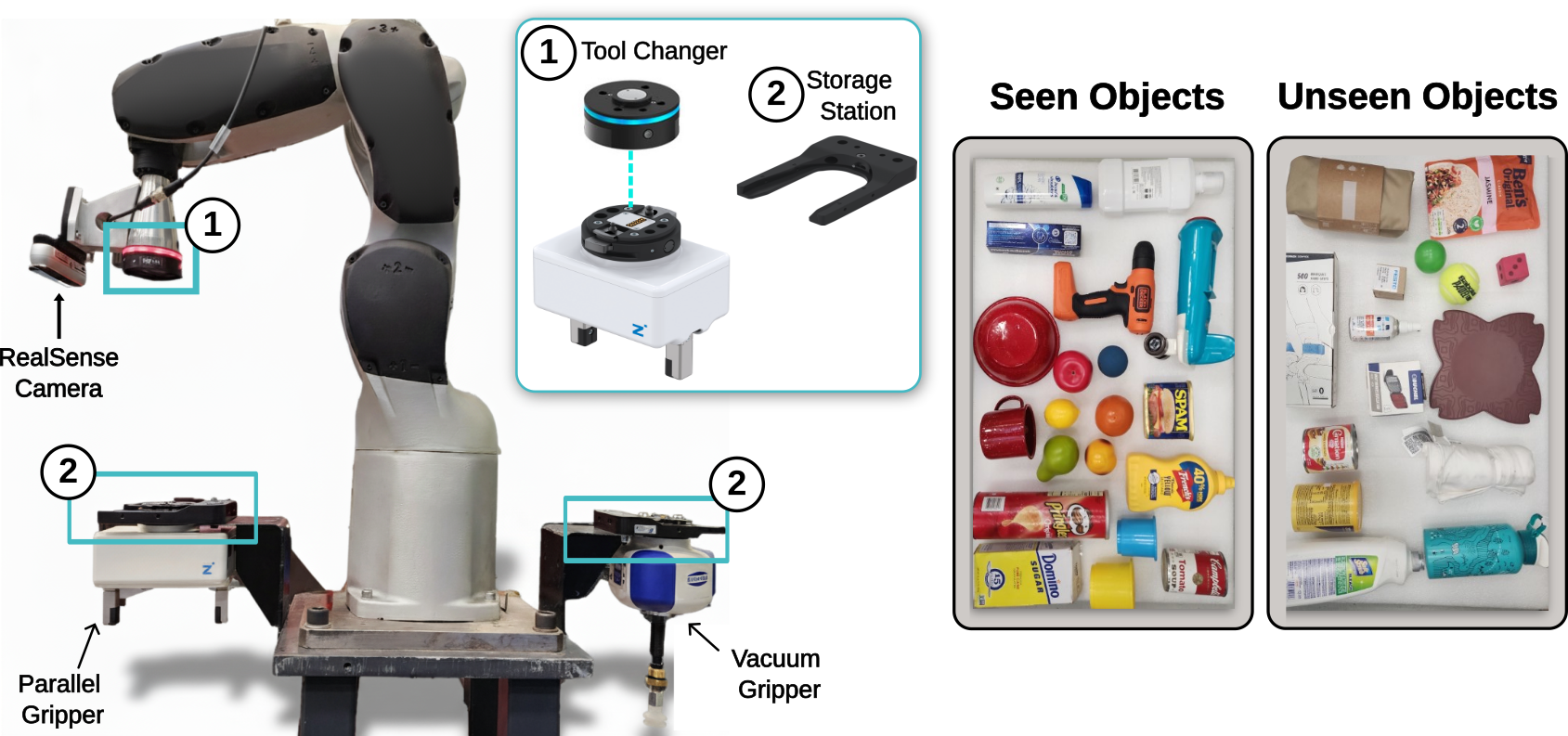}
\caption{\steph{Left:} Our \mg system with depth camera includes a parallel and vacuum gripper stored in dedicated stations (2). 
The automated tool changer (1) facilitates quick switching by connecting directly to the gripper and removing it from its station.
\steph{Right: Real-world test objects.}}
 \vspace{-2mm}
\label{fig_4}
\end{figure}

To conduct a comprehensive evaluation of our proposed approach, we perform real-world experiments in an industrial setting. Qualitative video demonstrations of the grasp executions are provided in the supplementary material.
All experiments are executed using a 6-DoF industrial robotic arm, equipped with a RealSense camera for 3D perception, as shown in Fig. \ref{fig_4}. The end-effector configuration consists of a parallel gripper and a vacuum gripper, coupled via an automated tool changer that enables their rapid switching during operation.

To quantify the performance and to understand whether failures are due to the quality of the grasping pose or perceptual limitations, we employ the following metrics:

\begin{itemize}
    \item $R_{object}$: ratio of the number of successfully cleared objects to the total number of objects.
    \item $R_{grasp}$ (success rate): ratio of the number of successful grasps to the total number of grasps.
    \item $R_{mix}$: ratio of the number of grasps (over the successfully cleared objects) to the number of successfully cleared objects.
    \item $R_{seen}$: ratio of the number of detected objects to the total number of objects in the scene.
\end{itemize}

We adopt the same experimental setup as previous works~\cite{shao2019suction, graspnet1b,wu2024economic}. For each trial, the system processes the input point cloud and predicts grasp candidates. The one with the highest score is executed.
After each attempt, the system re-evaluates the scene and repeats the process. The cycle continues until all objects are successfully cleared from the table or the robot fails to grasp any object for three consecutive times.

To assess performance in diverse cluttered environments, we consider ten test scenes belonging to the following categories:
\begin{itemize}

    \item \textit{Seen objects:} Five scenes are populated by randomly arranging six items from the seen objects (Fig. \ref{fig_4}). This setup, with a total of 30 object instances, ensures variability in shape, texture, and physical properties while including repetitions across scenes.

    \item \textit{Unseen objects:} Five scenes are composed using five randomly selected items from completely new objects. As shown in Fig. \ref{fig_4}, these items introduce novel shapes, colors, deformable materials \steph{and irregular surfaces} to evaluate the model's generalization capabilities across 25 total objects.

\end{itemize}

\begin{table}[t]
    \centering
    \caption{Real-world Experiments Performance.}
    \label{tab:RealWorld}
    \setlength\extrarowheight{2pt}
    \resizebox{\columnwidth}{!}{
    \begin{tabular}{Sc@{~~}Sc@{~~}Sc@{~~}Sc@{~~}Sc@{~~}Sc@{~~}Sc@{~~}Sc@{~~}Sc@{~~}}
        \toprule
        \multirow{2.5}{*}{\textbf{\small{Method}}} & \multicolumn{4}{c}{\textbf{\small{Seen Objects}}} & \multicolumn{4}{c}{\textbf{\small{Unseen Objects}}} \\
        \cmidrule(lr){2-5} \cmidrule(lr){6-9}& $R_{object}\uparrow$ & $R_{grasp}\uparrow$ & $R_{mix}\downarrow$ & $R_{seen}\uparrow$ & $R_{object}\uparrow$ & $R_{grasp}\uparrow$ & $R_{mix}\downarrow$ & $R_{seen}\uparrow$ \\
        \midrule
        & \multicolumn{8}{c}{\textbf{\smaller{Single-gripper}}}\\
        \midrule
        & \multicolumn{8}{c}{\textbf{\smaller{Vacuum}}}\\
        \cmidrule{2-9}
        S1B \cite{cao2021suctionnet} & \makecell{16$/$30 \\ (53\%)} & 39.11$\%$ & \textbf{1.14} & \makecell{24$/$30 \\ (80\%)} & \makecell{12$/$25 \\ (48\%)} & 43.33$\%$ & 1.17 & \makecell{17$/$25 \\ (68\%)} \\
        \rowcolor{gray!15}
        \ours & \makecell{\textbf{21$/$30} \\ \textbf{(70\%)}} & \textbf{41.02$\%$} & 1.34 &  \makecell{\textbf{30$/$30} \\ \textbf{(100\%)}} &  \makecell{\textbf{20$/$25} \\ \textbf{(80\%)}} & \textbf{54.76$\%$} & \textbf{1.14} &  \Gape[0pt][2pt]{\makecell{\textbf{25$/$25}\\ \textbf{(100\%)}}}   \\
        \midrule
        & \multicolumn{8}{c}{\textbf{\smaller{Parallel}}}\\
        \cmidrule{2-9}
        EFG \cite{wu2024economic} & \makecell{28$/$30 \\ (93\%)} & 66.03$\%$ & 1.45 & \makecell{\textbf{30$/$30} \\  \textbf{(100\%)}} & \makecell{22$/$25 \\ (88\%)} & \textbf{76.98$\%$} & \textbf{1.14} & \makecell{24$/$25 \\ (96\%)} \\
        
        \rowcolor{gray!15}
        \ours & \makecell{\textbf{29$/$30} \\ \textbf{(96\%)}} & \textbf{76.64$\%$} & \textbf{1.35} & \makecell{29$/$30 \\ (96\%)} & \makecell{\textbf{23$/$25} \\ \textbf{(92\%)}} & 70.48$\%$ & 1.16 & \Gape[0pt][2pt]{\makecell{\textbf{25$/$25} \\  \textbf{(100\%)}}} \\
%------------MULTIGRIPPER-----------------        
        \midrule
        & \multicolumn{8}{c}{\textbf{\smaller{\Mg with online switching}}}\\
        \midrule

% S1B+EFG with z-score
        \makecell{S1B + EFG \\ w/ z-score} & \makecell{22$/$30 \\ (73\%)} & 42.52\%  &  \textbf{1.32} &   \makecell{29$/$30 \\ (96\%)}   &  \makecell{18$/$25 \\ (72\%)} & 50.79\% &  1.40 & \makecell{23$/$25 \\ (92\%)}   \\

% S1B+EFG with Switching Strategy
        \makecell{S1B + EFG \\ w/ SS} & \makecell{ 23$/$30\\ (76.67\%)}  &  49.65\% & 1.48 & \makecell{ 30$/$30\\(100\%)}   & \makecell{ 16$/$25\\ (64\%)}  & \textbf{80\%} &  1.4 & \makecell{ 18$/$25\\ (72\%)}   \\

% MultiGraspNet with z-score
        \rowcolor{gray!15}
        \makecell{\ours \\ w/ z-score} & \makecell{ 25$/$30\\ (83.33\%) } &  57.23\% &  \textbf{1.32 }&   \makecell{ \textbf{30$/$30}\\ \textbf{(100\%)} }   &  \makecell{ 21$/$25\\ (84\%) }& 54.06\% &  1.38 & \makecell{ \textbf{25$/$25}\\ \textbf{(100\%)} }  \\

% MultiGraspNet with Switching Strategy
        \rowcolor{gray!15}
        \makecell{\ours \\ w/ SS} & \makecell{\textbf{29$/$30} \\ (\textbf{96\%})} & \textbf{70.83\%} & 1.35 &  \makecell{\textbf{30$/$30} \\ \textbf{(100\%)}} & \makecell{\textbf{24$/$25} \\ (\textbf{96\%})}  & 74.29 \% & \textbf{1.16} & \Gape[0pt][2pt]{\makecell{
        \textbf{25$/$25}\\ \textbf{(100\%)}}}\\
        \bottomrule
    \end{tabular}
    }
\end{table}

To establish a baseline, we compare \ours against two single-gripper methods, S1B \cite{cao2021suctionnet} for vacuum and EFG \cite{wu2024economic} for parallel grasping in two different modalities.

\subsubsection{Single-gripper}  For a direct comparison with single-gripper methods, we evaluate \ours by isolating the vacuum and parallel predictions against their respective single-gripper counterparts.
The results reported in Table~\ref{tab:RealWorld} indicate that our approach successfully grasps more objects with fewer attempts than S1B~\cite{cao2021suctionnet}, as reflected by the higher $R_{object}=70\%$ and $R_{grasp}=41.02\%$  for seen objects, 
and $R_{object}=80\%$ and $R_{grasp}=54.76\%$ for the unseen ones.
For single-parallel grasping, we obtain competitive results compared with EFG \cite{wu2024economic} for the seen objects by picking one more item ($R_{object}=96\%$) 
with fewer attempts ($R_{grasp}=76.64\%$). 
For the unseen objects, our method grasps one more object, 
although requiring slightly more attempts.

\subsubsection{Multi-gripper with online switching}

To analyze the performance of \ours including autonomous gripper selection and switching,
we compare our method using the proposed switching strategy (named \textit{\ours~w$/$~SS}) with the following baselines:

\begin{itemize}
    \item \textit{S1B+EFG w$/$ z-score:} combining S1B+EFG via z-score normalization;
    \item \textit{S1B+EFG w$/$ SS:} a version of S1B+EFG utilizing a dedicated module analogous to our switching strategy;
    \item\textit{\ours w$/$ z-score:} our method adapted with z-score normalization to ensure a fair evaluation.
\end{itemize}

Table \ref{tab:RealWorld} shows that S1B+EFG w$/$ z-score results in large performance reductions with respect to the single-task EFG baseline on both seen and unseen objects (e.g., $-20\%$ and $-16\%$ in terms of $R_{object}$, respectively). We observe a similar trend for \ours w$/$ z-score,  which successfully grasps markedly fewer objects than parallel-only \ours. These results show that simple normalization yields lower performance than single-gripper systems, highlighting that appropriate gripper selection is critical for \mg systems.

Finally, we evaluate \ours and S1B+EFG using our learning-based switching strategy for both methods. We notice that S1B+EFG w$/$ SS slightly improves in terms of $R_{object}$ for seen objects, while it decreases for unseen ones.
However, in both cases it still performs poorly with respect to single-gripper baselines. On the contrary, \ours~w$/$~SS achieves $R_{object}=96\%$ on both seen and unseen objects, matching or surpassing all baseline results, including S1B+EFG w$/$ SS. 

Overall, our experimental results show that incorporating shared features allows \ours to achieve a more \steph{complete} understanding of the scene compared to the isolated, single-task features of the baselines.
This results in higher performance on downstream multi-gripper grasping tasks.

\section{Conclusion}

We present \ours, a \mt 3D deep learning model capable of predicting grasp poses for both parallel-jaw and vacuum grippers within a unified framework.
By leveraging complementary information 
between grasping modalities, our model generates accurate grasp poses across a variety of cluttered scenes with diverse objects.
Furthermore, our \mg experiments with online switching highlight the advantages of our unified architecture, significantly outperforming traditional normalization-based \mg approaches and combined single-task baselines in real-world scenarios. 
By integrating multiple modalities in a shared architecture, \ours provides an accurate, efficient, and versatile solution for real-world robotic grasping.
\steph{Future work will investigate the integration of a supervision-free switching strategy alongside a learning-based module to minimize
the number of gripper switches while accounting for picking ordering and collisions.}

%\appendix
%\input{Sections/6_Appendix}

\bibliography{references}
\bibliographystyle{IEEEtran}

\end{document}